\documentclass[sigconf]{acmart}

\usepackage{amsmath,graphicx}

\usepackage{amssymb}
\usepackage{amsfonts}
\usepackage{url}
\usepackage{algorithm}
\usepackage{algorithmicx}
\usepackage{algpseudocode}
\usepackage{booktabs}
\usepackage{threeparttable}
\usepackage{multirow}
\usepackage{subcaption}
\usepackage{caption}
\usepackage{array}
\usepackage{color}
\usepackage{soul}

\AtBeginDocument{%
  \providecommand\BibTeX{{%
    \normalfont B\kern-0.5em{\scshape i\kern-0.25em b}\kern-0.8em\TeX}}}

\setcopyright{acmcopyright}
\copyrightyear{2021}
\acmYear{2021}

\acmConference[]{KDD Workshop on Model Mining}{2021}{}
\acmBooktitle{KDD Workshop on Model Mining}
\acmPrice{15.00}
\acmISBN{978-1-4503-XXXX-X/18/06}

\begin{document}

\title{Structured DropConnect for Uncertainty Inference in Image Classification}


\author{Wenqing Zheng$^1$, Jiyang Xie$^1$, Weidong Liu$^2$, Zhanyu Ma$^1$,$^3$}
\affiliation{
$^1$ Pattern Recognition and Intelligent System Lab., Beijing University of Posts and Telecommunications, Beijing, China\\
$^2$ China Mobile Research Institute, Beijing, China\\
$^3$ Beijing Academy of Artificial Intelligence}
 


\begin{abstract}
With the complexity of the network structure, uncertainty inference has become an important task to improve classification accuracy for artificial intelligence systems. For image classification tasks, we propose a structured DropConnect (SDC) framework to model the output of a deep neural network by a Dirichlet distribution. We introduce a DropConnect strategy on weights in the fully connected layers during training. In test, we split the network into several sub-networks, and then model the Dirichlet distribution by match its moments with the mean and variance of the outputs of these sub-networks. The entropy of the estimated Dirichlet distribution is finally utilized for uncertainty inference. In this paper, this framework is implemented on LeNet$5$ and VGG$16$ models for misclassification detection and out-of-distribution detection on MNIST and CIFAR-$10$ datasets. Experimental results show that the performance of the proposed SDC can be comparable to other uncertainty inference methods. Furthermore, the SDC is adapted well to different network structures with certain generalization capabilities and research prospects.
\end{abstract}



\keywords{Uncertainty inference, image classification, DropConnect, Dirichlet distribution}

\maketitle

\section{Introduction}

As a foundation of computer vision (CV), image classification also supports many other important tasks, such as object detection, object tracking, and image segmentation~\cite{karpathy2015deep}. Although the image classification task is simple for humans, it is still challenging for automatic systems, due to overfitting and overconfidence of deep neural networks (DNNs). Therefore, for a long period, the improvement of classification accuracy by increasing the robustness of a CV system is one of the key contents of research. 
However, the internal process of the DNNs is a black box to us~\cite{eldesokey2020uncertainty}, which means that we cannot know how the network infers the final prediction results, nor whether the output prediction results can be trusted. Huang et al.~\cite{huang2020universal} demonstrated that it is possible to use physical and digital operations to deceive the most advanced object detectors and make them produce false but highly deterministic predictions. Similar statements can be also found in~\cite{goodfellow2014explaining}. Therefore, We hope that the network can tell us whether the predictions are confident, or called uncertainty, instead of only obtaining prediction values~\cite{goodfellow2014explaining}. In the applications with high error costs such as autonomous driving and medical diagnosis, we are eager to obtain the uncertainty of models' prediction results, so that we can prevent catastrophic accidents in time and ensure security~\cite{amodei2016concrete} of the intelligent systems. 

\begin{figure}[!t]
  \centering
  \includegraphics[width=\linewidth]{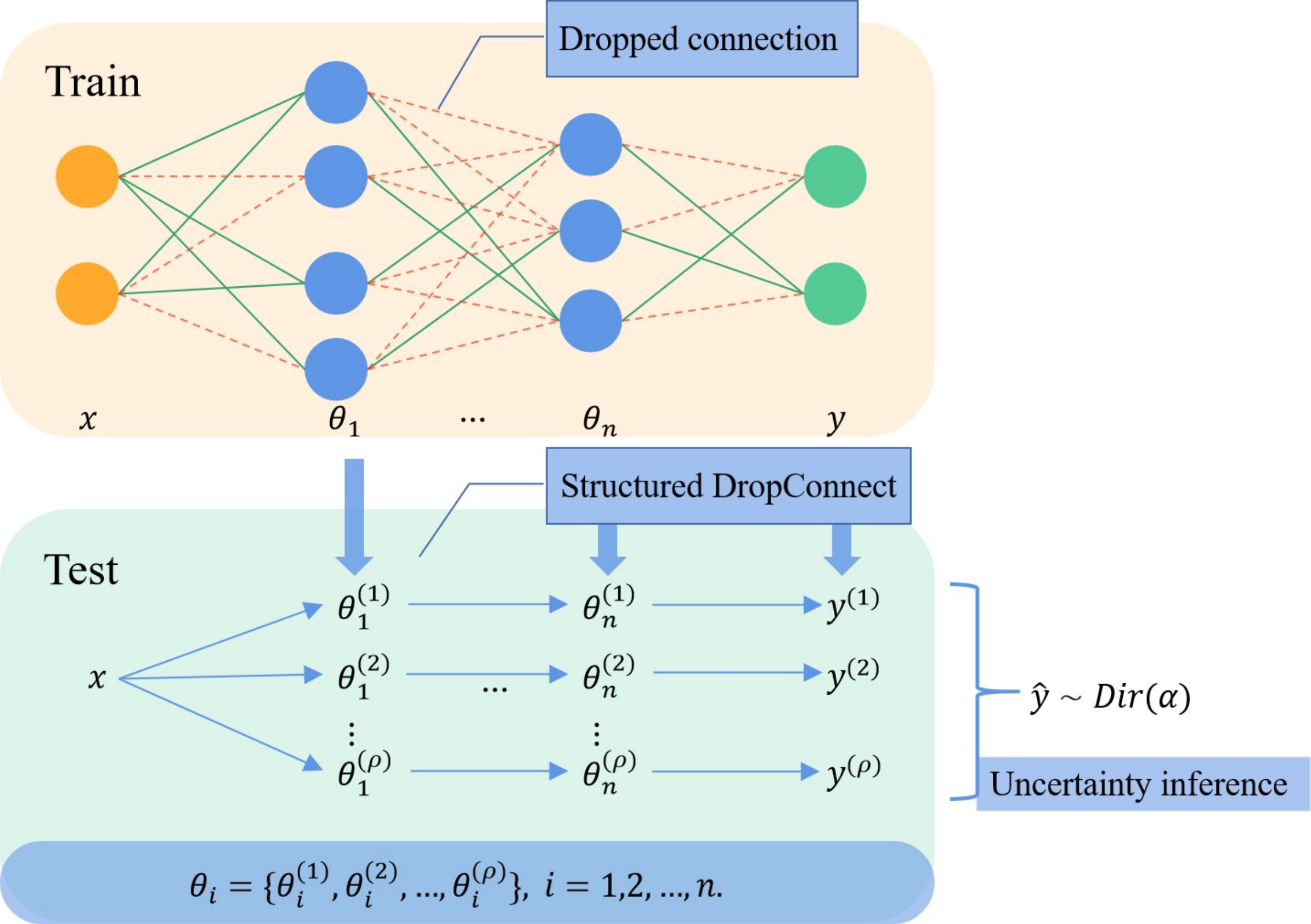}
  \caption{Illustration of the proposed structured DropConnect (SDC). In train phase, DropConnect is used on the model weights of the fully connected (FC) layers (instead of hidden nodes as in dropout) with dropout rate as $1/\rho$. While in test stage, the model weights in the network are splited into $\rho$ distinct groups and the final softmax output result can be considered a series of samples from a Dirichlet distribution.}\label{fig:structure}
\end{figure}

The methods of uncertainty inference in deep learning mainly focus on three aspects, including model architecture~\cite{malinin2018predictive}, quantification or evaluation criteria~\cite{wan2013regularization}, and data enhancement~\cite{wang2019aleatoric}. In terms of model architecture, deep ensemble~\cite{lakshminarayanan2016simple} estimates the uncertainty of the model by training multiple models and calculating the variance of their output predictions. Gal and Ghahramani~\cite{gal2016dropout} proposed to use Monte Carlo dropout (MC-dropout) to estimate predictive uncertainty by using dropout~\cite{srivastava2014dropout} at test time. Interestingly, dropout can be also interpreted as ensemble model combination~\cite{srivastava2014dropout}, where the predictions are averaged over an ensemble of DNNs with parameter sharing. The majority methods of uncertainty inference revolves around a Bayesian formalism, which is computationally expensive. So a series of Bayesian approximation methods have been proposed. Among them, the uncertainty is mainly used in misclassification detection and out-of-distribution detection. 

Inspired by the deep ensemble and MC-dropout, we propose a new strategy, namely structured DropConnect (SDC) in Figure~\ref{fig:structure} to model the output distribution of a network. The network is divided into two processes,~\emph{i.e.}, training and test. In training phase, DropConnect is used on the model weights of fully connected (FC) layers (instead of hidden nodes as in dropout) with dropout rate as $\frac{1}{\rho}$ where $\rho$ is a positive integer. While in test phase, the weights in the network are splited into $\rho$ groups, so that the final softmax output result can be considered as a series of samples from a Dirichlet distribution. The parameters of the Dirichlet distribution can be estimated by matching their moments with the mean and variance of the output value. The uncertainty of the network can be modeled according to the entropy of the Dirichlet distribution.

The contributions of this paper is three-fold: 

\begin{itemize}

\item A SDC framework is proposed. We implement structure DropConnect on the model weights of the FC layers during test to obtain a set of sub-networks, we also estimate a Dirichlet distribution by the outputs of the sub-networks for uncertainty inference.

\item The proposed SDC is implemented in two common backbones including LeNet$5$ and VGG$16$ models. 

\item We evaluate the propsoed SDC on misclassification detection and out-of-distribution (OOD) detection tasks on MNIST and CIFAR-$10$ datasets, respectively. The experimental results are competitive with previous uncertainty inference methods.

\end{itemize}

\section{Related Work}

Uncertainty can be roughly classified into two categories,~\emph{i.e.}, epistemic and aleatoric uncertainty~\cite{senge2014reliable}. The former captures the uncertainty in model parameters caused by the lack of training data. The latter mainly comes from noises of data. There is usually a third source called distributional uncertainty, which is generally caused by the mismatch in the distribution of training and test data, which is also a common situation in practical applications. 

Most of studies~\cite{mackay1992practical,mackay1992bayesian,hinton1993keeping} in uncertainty inference were inspired by Bayesian statistics. The Bayesian neural network (BNN)~\cite{neal2012bayesian} is a probabilistic version of the traditional neural network that has a prior distribution on each weight. The essence of this network is suitable for generating uncertainty estimates. Since exact Bayesian inference is computationally intractable for DNNs, the approximation of the BNN is a hot topic of current research, including deterministic approaches~\cite{degroot1983comparison}, Markov chain Monte Carlo with Hamiltonian dynamics~\cite{neal1993bayesian}, and variational inference~\cite{gal2016dropout}. 
Most of the recently proposed variational inference-based methods usually focused on approximating a distribution on the output of a DNN, such as Gaussian distribution, softmax distribution, or Dirichlet distribution, and calculated the mean values of the output distribution to quantify uncertainty. Blundell et al.~\cite{blundell2015weight} proposed a backpropagation compatible algorithm with unbiased MC gradient to estimate the parameter uncertainty of DNNs, which is called Bayesian by Backpropagation (BBP). Hendrycks and Gimpe~\cite{hendrycks2016baseline} introduced a baseline model that assumes the outputs follow the softmax distribution and detected misclassification of distribution samples with the largest softmax in multiple tasks. In Dirichlet prior network (DPN)~\cite{malinin2018predictive}, a Dirichlet distribution was introduced into the DNNs for modeling uncertainty. Xie~\cite{9439951} proposed advanced dropout, a model-free methodology, to mitigate overfitting and improve the performance of DNNs.

\section{METHODOLOGY}

\subsection{Dirichlet Distribution}

Inspired by the Bayesian learning-based method~\cite{malinin2018predictive}, uncertainty inference can be understood as modeling the output predictions of the network. Here, the Dirichlet distribution is selected and estimated for output predictions of the network. We define the positive parameters of the Dirichlet distribution as $\alpha=[\alpha_1,\cdots,\alpha_K]^{\text{T}}$. $\alpha_0$ is the sum of all $\alpha_i$, called the accuracy of the Dirichlet distribution. The higher the value $\alpha_0$ is, the sharper the Dirichlet distribution is. The probability density function (PDF) of a Dirichlet distributed vector $\mu$ is
\begin{equation}
  p(\mu|\alpha)=\frac{\Gamma(\alpha_0)}{\prod_{i=1}^K \Gamma(\alpha_i)}\prod_{i=1}^K\mu_i^{\alpha_i-1},\ \alpha_i>0,\ \alpha_0=\sum_{i=1}^K\alpha_i,
\end{equation}
where $\sum_{i=1}^K\mu_i=1,\mu_i>0,i=1,\cdots,K$, the random variable $\mu$ obeys the Dirichlet distribution with parameter $\alpha$, denoted as $\mu\sim Dir(\alpha)$.  

Supposing a group of $N$ vector samples $\{\mu^{(1)},\cdots,\mu^{(N)}\}$ that follows the Dirichlet distribution $Dir(\alpha)$, the mean and variance of the Dirichlet distribution can be matched by those of the samples~\cite{xie2018balson} as
\begin{align}
  E[\mu_i]&=\frac{\alpha_i}{\alpha_0},\label{eq:mean}\\ Var[\mu_i]&=\frac{\alpha_i(\alpha_0-\alpha_i)}{\alpha_0^2(\alpha_0+1)}.\label{eq:var}
\end{align}
Thus, we can deduce the parameters of the Dirichlet distribution by solving the aforementioned equations as
\begin{equation}
  \alpha_i=E[\mu_i](\frac{E[\mu_i](1-E[\mu_i])}{Var[\mu_i]}-1).\label{eq:estimatealpha}
\end{equation}

\begin{table*}
  \caption{Performance of the misclassification detection on MNIST and CIFAR-$10$ datasets. We take AUROC (\%) as the evaluation criterion.}
  \label{tab:miscls}
  \begin{tabular}{lcccccc}
    \toprule
    Dataset & \multicolumn{3}{c|}{MNIST} & \multicolumn{3}{c}{CIFAR-10}\\
    \midrule
    \texttt Method & Max.P & Ent. & \multicolumn{1}{c|}{D.Ent} & Max.P & Ent. & D.Ent \\
    \midrule
    \texttt Baseline & $97.52\pm0.23$ & $97.38\pm0.25$ & -  &$91.30\pm0.37$&$91.30\pm0.37$&-\\
    \texttt MC Dropout & $97.74\pm0.45$ & $97.64\pm0.46$ & -  &$89.54\pm0.24$&$89.42\pm0.23$&-\\
    \texttt SDC (ours) & $97.75\pm0.22$ & $97.74\pm0.22$ & $94.84\pm0.49$  &$87.92\pm0.43$&$89.72\pm0.42$&$85.92\pm0.80$\\
    \bottomrule
  \end{tabular}
\end{table*}

\subsection{DropConnect}

DropConnect~\cite{wan2013regularization} is a generalization of Dropout~\cite{gal2016dropout}, where the Bernoulli distributed dropout masks are applied directly to each model weight of the network instead of hidden nodes. When applying the DropConnect to the model weights of an FC layer, the activation function of the layer output can be written as
\begin{equation}
    y_l=\sigma\left((Z_l\otimes W_l)x_l\right),
\end{equation}
where $x_l$ and $y_l$ are input and output features of the $l^{th}$ layer, respectively, $\sigma(\cdot)$ is activation function, $Z_l$ is the binary matrix with the same shape as model weights $W_l$, whose elements follow as
\begin{align}
  Z_l^{(i,j)}&=Bernoulli(p_l),\\
  l&=1,\cdots,L,\ i=1,\cdots,h_{l},\ j=1,\cdots,h_{l-1},\ \nonumber
\end{align}
where $z_l^{(i,j)}$ is a random binary value, which is associated with the model weight from the $j^{th}$ hidden node of the $(l-1)^{th}$ layer to the $i^{th}$ hidden node of the $l^{th}$ layer, $h_{l}$ is hidden node number of the $l^{th}$ layer. $z_l^{(i,j)}$ takes a value of $1$ with probability $p$ (assuming that all weights in a layer have the same probability value). $p_l=1-\frac{1}{\rho_l}$ with positive integer $\rho_l$.

\subsection{Structured DropConnect}

As mentioned above, in order to model the output distribution of the network to quantify the uncertainty, we propose a framework called Structured DropConnect (SDC) to estimate a Dirichlet distribution (Figure~\ref{fig:structure}). Among them, the network can be intuitively considered as spliting the network into a set of sub-networks during test, so as to obtain a set of output samples rather than a single output prediction. The mean and variance of the output samples can be used to estimate the parameters of Dirichlet distribution. 

Specifically, we initialize a set of masks for each layer under a structured condition, each of which is a binary matrix with the same size of the model weights. We multiply the masks onto the corresponding model weight matrices to obtain the splited weights of the sub-networks. The zero value in the masks represents that the element in the weight matrices is discarded. It is worth noting that we define the structured condition that the ratio of zero elements in each mask matrix in a set of masks is set equal to $\frac{1}{\rho}$. Meanwhile, the positions of zero elements in different sub-networks are absolutely different from each other, which ensures the diversity of network split. Furthermore, In order to correspond to test phase, dropout rate of each layer is also set as $\frac{1}{\rho}$ during training.

Correspondingly, in test stage, the number of sub-networks obtained by the network split is as $\rho$ (for simplicity of derivation, the value is an integer) and the final output predictions for one input are $\rho$ folds, which is defined as the $\rho$ samples. The Dirichlet distribution parameters can be then estimated by~\eqref{eq:estimatealpha}, and the uncertainty of the network corresponding to the input can be quantified according to the entropy of the Dirichlet distribution. Applying the SDC, the output activation function can be written as
\begin{equation}
    y_l^{(m)}=\sigma\left((Mask^{(m)}_l\otimes W_l)x_l\right), m=1,\ \cdots,\rho,
\end{equation}
where $Mask^{(m)}_l$ indicates the mask of the $m^{th}$ split of the $l^{th}$ layer.

In particular, when the parameters of the Dirichlet distribution are estimated by a set of sub-networks, it can be known from the nature of the Dirichlet distribution that if $0<\alpha_i<1$, the Dirichlet distribution is highly sparse in dimension $i$, where $K$ indicates the number of $\alpha_i,\ i=1,\cdots,K$, while it is dense if $\alpha_i\ge1$. Thus, $\alpha_i$ obtained by~\eqref{eq:estimatealpha} is further processed by
\begin{equation}
    \alpha_i \leftarrow
\begin{cases} 
    \frac{\alpha_i-1}{\sum_{j|\alpha_i>1} \alpha_j -K},  & \alpha_i>1 \\
    0, & otherwise
\end{cases}.
\end{equation}

\begin{table*}
  \caption{Performance of out-of-distribution (OOD) detection on MNIST and CIFAR-$10$ datasets. We take AUROC (\%) as the evaluation criterion.}
  \label{tab:ood}
  \begin{tabular}{lcccccc}
    \toprule
    Dataset & \multicolumn{3}{c|}{MNIST} & \multicolumn{3}{c}{CIFAR-10}\\
    \midrule
    \texttt Method & Max.P & Ent. & \multicolumn{1}{c|}{D.Ent} & Max.P & Ent. & D.Ent \\
    \midrule
    \texttt Baseline & $97.52\pm0.24$ & $98.12\pm0.22$ & -  &$87.50\pm0.41$ & $88.70\pm0.39$ &-\\
    \texttt MC Dropout & $98.24\pm0.35$ & $98.08\pm0.36$ & -  &$87.60\pm0.52$ & $89.20\pm0.33$ &-\\
    \texttt SDC(ours) & $98.24\pm0.32$ & $98.19\pm0.32$ &  $94.21\pm0.71$ &$92.52\pm0.51$ & $90.36\pm0.46$ & $88.44\pm0.72$\\
    \bottomrule
  \end{tabular}
\end{table*}

\section{Experimental Results and Discussions}

We evaluated the proposed SDC in two uncertainty inference tasks including misclassification detection and out-of-distribution (OOD) detection. The experiments were conducted on MNIST~\cite{lecun1990handwritten} and CIFAR-$10$~\cite{krizhevsky2009learning} datasets with LeNet$5$~\cite{lecun1990handwritten} and VGG$16$~\cite{simonyan2014very} as backbones, respectively, to assess the SDC’s ability in uncertainty inference. Both backbones are composed of several convolutional, pooling, and FC layers, while we only implemented the proposed SDC in the FC layers. Following the settings in~\cite{hendrycks2016baseline}, we introduced max probability (Max.P.) and entropy (Ent.) of output probabilities as uncertainty measurement metrics. We also added differential entropy (D.Ent.) for the proposed SDC as an extra metric. We utilized the area under receiver operating characteristic curve (AUROC) to evaluate the performance of uncertainty inference. We compared the proposed SDC with the existing uncertainty inference algorithm MC-Dropout~\cite{gal2016dropout} and the baseline~\cite{hendrycks2016baseline}.

We trained $40$ epochs for the MNIST dataset, and $100$ epochs for the CIFAR-$10$ dataset, with batch size $256$ for each dataset. The dropout rate was $0.1$, that is, the network was divided into $10$ sub-networks during test. The initialized learning rate was set as $7.5\times10^{-4}$. Cross-entropy loss was used as the loss function during training and Adam~\cite{kingma2014adam} was used as the optimizer. We ran each network for $5$ times to obtain the mean and variance of the uncertainty metrics. 


\subsection{Misclassification Detection}

An important task in uncertainty inference is misclassification detection, which is dedicated to detecting misclassified samples in test sets with uncertainty. Therefore, after obtaining the well-trained model, we follow the previously proposed uncertainty quantification method to give an uncertainty inference result for different inputs in test set and detect whether an input is misclassified. Among them, the misclassified predictions represent positive samples. 

 The experimental performance on the MNIST and the CIFAR-$10$ datasets is shown in Table~\ref{tab:miscls}. Among them, on the MNIST dataset, the AUROC indicator of the proposed SDC is the highest value and the performance of both the mean and the variance is the best (the variance is the smaller the better). On the CIFAR-$10$ dataset, the proposed SDC ranks second best under the AUROC of Ent. In terms of overall performance on the CIFAR-$10$, the proposed SDC is basically comparable to the other two uncertainty inference methods in the experiments.

\subsection{Out-of-Distribution Detection}

Out-of-distribution (OOD) detection task is similar to the former one that finds abnormal samples. The difference is that the OOD detection aims to find a test sample whether belongs to the training set. In the OOD detection, test sets of street view house numbers (SVHN)~\cite{netzer2011reading} and tiny ImageNet (TIM)~\cite{russakovsky2015imagenet} datasets were used as the OOD data for the MNIST and the CIFAR-$10$ datasets, respectively. 

The results are shown in Table~\ref{tab:ood}. It can be seen that in the comparisons of the AUROC, the proposed SDC is superior to the other two uncertainty inference methods in terms of performance and can well detect the OOD data on both datasets. In general, the performance of the proposed SDC in the OOD detection is quite satisfactory and it has application potential.

\subsection{Discussion}

Through the two datasets and the two backbone models, we have obtained the evaluation results in the misclassification detection and the OOD detection. Overall, the proposed SDC has a competitive performance in uncertainty inference. According to the characteristics of the SDC and the experimental settings, the following three points of view are worth to be discussed: 

\begin{itemize}

\item Through the aforementioned experiments, the integration of several sub-networks can actually improve the performance of the uncertainty inference. However, we have no idea with which factor(s) can effect the performance improvement. The possible factors include the way of network split, the number of network splits, the choice of output distribution,~\emph{etc}. Different factors may all work on the final results. 

\item The evaluation criteria of uncertainty quantification are closely related to the experimental results. The proposed SDC only provides a feasible way for uncertainty quantification. However, to our best knowledge, there are no stable standards for quantitative uncertainty metric and criteria in the researches of uncertainty. 

\item In view of the extensive applications based on image classification, there are many related datasets and network structures. This paper is committed to giving a solution with strong generalization abilities. However, generalization ability is related to many factors. Future research can be devoted to the improvement of it, and it is expected to make more contributions in uncertainty inference or cross-domain learning. 

\end{itemize}

\section{Conclusion}

In this paper, a new framework called Structured DropConnect (SDC) was proposed to model output predictions by using Dirichlet distribution. In test phase, the model weights of the FC layers were splited to obtain a group of sub-networks. Thus, the predicted value of the original DNN output was modified into a group of predicted values. The mean and variance of the predicted values were calculated, so as to estimate the Dirichlet distribution. We used the AUROC to evaluate the performance of uncertainty quantification. The proposed SDC was evaluated on the MNIST and the CIFAR-$10$ datasets with LeNet$5$ and VGG$16$ models as backbone models. The experimental results of the proposed SDC in the misclassification detection and the OOD detection were compared with those of the baseline and the MC-dropout method. The results showed that our method improved performance on uncertainty inference. The proposed SDC can also be used in other tasks and networks, such as object detection and semantic segmentation. 


\bibliographystyle{ACM-Reference-Format}
\bibliography{SDC}


\begin{thebibliography}{28}


\ifx \showCODEN    \undefined \def \showCODEN     #1{\unskip}     \fi
\ifx \showDOI      \undefined \def \showDOI       #1{#1}\fi
\ifx \showISBNx    \undefined \def \showISBNx     #1{\unskip}     \fi
\ifx \showISBNxiii \undefined \def \showISBNxiii  #1{\unskip}     \fi
\ifx \showISSN     \undefined \def \showISSN      #1{\unskip}     \fi
\ifx \showLCCN     \undefined \def \showLCCN      #1{\unskip}     \fi
\ifx \shownote     \undefined \def \shownote      #1{#1}          \fi
\ifx \showarticletitle \undefined \def \showarticletitle #1{#1}   \fi
\ifx \showURL      \undefined \def \showURL       {\relax}        \fi
\providecommand\bibfield[2]{#2}
\providecommand\bibinfo[2]{#2}
\providecommand\natexlab[1]{#1}
\providecommand\showeprint[2][]{arXiv:#2}

\bibitem[\protect\citeauthoryear{Amodei, Olah, Steinhardt, Christiano,
  Schulman, and Man{\'e}}{Amodei et~al\mbox{.}}{2016}]%
        {amodei2016concrete}
\bibfield{author}{\bibinfo{person}{Dario Amodei}, \bibinfo{person}{Chris Olah},
  \bibinfo{person}{Jacob Steinhardt}, \bibinfo{person}{Paul Christiano},
  \bibinfo{person}{John Schulman}, {and} \bibinfo{person}{Dan Man{\'e}}.}
  \bibinfo{year}{2016}\natexlab{}.
\newblock \showarticletitle{Concrete problems in AI safety}.
\newblock \bibinfo{journal}{\emph{arXiv preprint arXiv:1606.06565}}
  (\bibinfo{year}{2016}).
\newblock


\bibitem[\protect\citeauthoryear{Blundell, Cornebise, Kavukcuoglu, and
  Wierstra}{Blundell et~al\mbox{.}}{2015}]%
        {blundell2015weight}
\bibfield{author}{\bibinfo{person}{Charles Blundell}, \bibinfo{person}{Julien
  Cornebise}, \bibinfo{person}{Koray Kavukcuoglu}, {and} \bibinfo{person}{Daan
  Wierstra}.} \bibinfo{year}{2015}\natexlab{}.
\newblock \showarticletitle{Weight uncertainty in neural network}. In
  \bibinfo{booktitle}{\emph{International Conference on Machine Learning}}.
  PMLR, \bibinfo{pages}{1613--1622}.
\newblock


\bibitem[\protect\citeauthoryear{DeGroot and Fienberg}{DeGroot and
  Fienberg}{1983}]%
        {degroot1983comparison}
\bibfield{author}{\bibinfo{person}{Morris~H DeGroot} {and}
  \bibinfo{person}{Stephen~E Fienberg}.} \bibinfo{year}{1983}\natexlab{}.
\newblock \showarticletitle{The comparison and evaluation of forecasters}.
\newblock \bibinfo{journal}{\emph{Journal of the Royal Statistical Society:
  Series D (The Statistician)}} \bibinfo{volume}{32}, \bibinfo{number}{1-2}
  (\bibinfo{year}{1983}), \bibinfo{pages}{12--22}.
\newblock


\bibitem[\protect\citeauthoryear{Eldesokey, Felsberg, Holmquist, and
  Persson}{Eldesokey et~al\mbox{.}}{2020}]%
        {eldesokey2020uncertainty}
\bibfield{author}{\bibinfo{person}{Abdelrahman Eldesokey},
  \bibinfo{person}{Michael Felsberg}, \bibinfo{person}{Karl Holmquist}, {and}
  \bibinfo{person}{Michael Persson}.} \bibinfo{year}{2020}\natexlab{}.
\newblock \showarticletitle{Uncertainty-aware cnns for depth completion:
  Uncertainty from beginning to end}. In \bibinfo{booktitle}{\emph{Proceedings
  of the IEEE/CVF Conference on Computer Vision and Pattern Recognition}}.
  \bibinfo{pages}{12014--12023}.
\newblock


\bibitem[\protect\citeauthoryear{Gal and Ghahramani}{Gal and
  Ghahramani}{2016}]%
        {gal2016dropout}
\bibfield{author}{\bibinfo{person}{Yarin Gal} {and} \bibinfo{person}{Zoubin
  Ghahramani}.} \bibinfo{year}{2016}\natexlab{}.
\newblock \showarticletitle{Dropout as a bayesian approximation: Representing
  model uncertainty in deep learning}. In
  \bibinfo{booktitle}{\emph{international conference on machine learning}}.
  PMLR, \bibinfo{pages}{1050--1059}.
\newblock


\bibitem[\protect\citeauthoryear{Goodfellow, Shlens, and Szegedy}{Goodfellow
  et~al\mbox{.}}{2014}]%
        {goodfellow2014explaining}
\bibfield{author}{\bibinfo{person}{Ian~J Goodfellow}, \bibinfo{person}{Jonathon
  Shlens}, {and} \bibinfo{person}{Christian Szegedy}.}
  \bibinfo{year}{2014}\natexlab{}.
\newblock \showarticletitle{Explaining and harnessing adversarial examples}.
\newblock \bibinfo{journal}{\emph{arXiv preprint arXiv:1412.6572}}
  (\bibinfo{year}{2014}).
\newblock


\bibitem[\protect\citeauthoryear{Hendrycks and Gimpel}{Hendrycks and
  Gimpel}{2016}]%
        {hendrycks2016baseline}
\bibfield{author}{\bibinfo{person}{Dan Hendrycks} {and} \bibinfo{person}{Kevin
  Gimpel}.} \bibinfo{year}{2016}\natexlab{}.
\newblock \showarticletitle{A baseline for detecting misclassified and
  out-of-distribution examples in neural networks}.
\newblock \bibinfo{journal}{\emph{arXiv preprint arXiv:1610.02136}}
  (\bibinfo{year}{2016}).
\newblock


\bibitem[\protect\citeauthoryear{Hinton and Van~Camp}{Hinton and
  Van~Camp}{1993}]%
        {hinton1993keeping}
\bibfield{author}{\bibinfo{person}{Geoffrey~E Hinton} {and}
  \bibinfo{person}{Drew Van~Camp}.} \bibinfo{year}{1993}\natexlab{}.
\newblock \showarticletitle{Keeping the neural networks simple by minimizing
  the description length of the weights}. In
  \bibinfo{booktitle}{\emph{Proceedings of the sixth annual conference on
  Computational learning theory}}. \bibinfo{pages}{5--13}.
\newblock


\bibitem[\protect\citeauthoryear{Huang, Gao, Zhou, Xie, Yuille, Zou, and
  Liu}{Huang et~al\mbox{.}}{2020}]%
        {huang2020universal}
\bibfield{author}{\bibinfo{person}{Lifeng Huang}, \bibinfo{person}{Chengying
  Gao}, \bibinfo{person}{Yuyin Zhou}, \bibinfo{person}{Cihang Xie},
  \bibinfo{person}{Alan~L Yuille}, \bibinfo{person}{Changqing Zou}, {and}
  \bibinfo{person}{Ning Liu}.} \bibinfo{year}{2020}\natexlab{}.
\newblock \showarticletitle{Universal physical camouflage attacks on object
  detectors}. In \bibinfo{booktitle}{\emph{Proceedings of the IEEE/CVF
  Conference on Computer Vision and Pattern Recognition}}.
  \bibinfo{pages}{720--729}.
\newblock


\bibitem[\protect\citeauthoryear{Karpathy and Fei-Fei}{Karpathy and
  Fei-Fei}{2015}]%
        {karpathy2015deep}
\bibfield{author}{\bibinfo{person}{Andrej Karpathy} {and} \bibinfo{person}{Li
  Fei-Fei}.} \bibinfo{year}{2015}\natexlab{}.
\newblock \showarticletitle{Deep visual-semantic alignments for generating
  image descriptions}. In \bibinfo{booktitle}{\emph{Proceedings of the IEEE
  conference on computer vision and pattern recognition}}.
  \bibinfo{pages}{3128--3137}.
\newblock


\bibitem[\protect\citeauthoryear{Kingma and Ba}{Kingma and Ba}{2014}]%
        {kingma2014adam}
\bibfield{author}{\bibinfo{person}{Diederik~P Kingma} {and}
  \bibinfo{person}{Jimmy Ba}.} \bibinfo{year}{2014}\natexlab{}.
\newblock \showarticletitle{Adam: A method for stochastic optimization}.
\newblock \bibinfo{journal}{\emph{arXiv preprint arXiv:1412.6980}}
  (\bibinfo{year}{2014}).
\newblock


\bibitem[\protect\citeauthoryear{Krizhevsky, Hinton, et~al\mbox{.}}{Krizhevsky
  et~al\mbox{.}}{2009}]%
        {krizhevsky2009learning}
\bibfield{author}{\bibinfo{person}{Alex Krizhevsky}, \bibinfo{person}{Geoffrey
  Hinton}, {et~al\mbox{.}}} \bibinfo{year}{2009}\natexlab{}.
\newblock \showarticletitle{Learning multiple layers of features from tiny
  images}.
\newblock  (\bibinfo{year}{2009}).
\newblock


\bibitem[\protect\citeauthoryear{Lakshminarayanan, Pritzel, and
  Blundell}{Lakshminarayanan et~al\mbox{.}}{2016}]%
        {lakshminarayanan2016simple}
\bibfield{author}{\bibinfo{person}{Balaji Lakshminarayanan},
  \bibinfo{person}{Alexander Pritzel}, {and} \bibinfo{person}{Charles
  Blundell}.} \bibinfo{year}{2016}\natexlab{}.
\newblock \showarticletitle{Simple and scalable predictive uncertainty
  estimation using deep ensembles}.
\newblock \bibinfo{journal}{\emph{arXiv preprint arXiv:1612.01474}}
  (\bibinfo{year}{2016}).
\newblock


\bibitem[\protect\citeauthoryear{LeCun, Boser, Denker, Henderson, Howard,
  Hubbard, and Jackel}{LeCun et~al\mbox{.}}{1990}]%
        {lecun1990handwritten}
\bibfield{author}{\bibinfo{person}{Yann LeCun}, \bibinfo{person}{Bernhard~E
  Boser}, \bibinfo{person}{John~S Denker}, \bibinfo{person}{Donnie Henderson},
  \bibinfo{person}{Richard~E Howard}, \bibinfo{person}{Wayne~E Hubbard}, {and}
  \bibinfo{person}{Lawrence~D Jackel}.} \bibinfo{year}{1990}\natexlab{}.
\newblock \showarticletitle{Handwritten digit recognition with a
  back-propagation network}. In \bibinfo{booktitle}{\emph{Advances in neural
  information processing systems}}. \bibinfo{pages}{396--404}.
\newblock


\bibitem[\protect\citeauthoryear{MacKay}{MacKay}{1992a}]%
        {mackay1992bayesian}
\bibfield{author}{\bibinfo{person}{David~JC MacKay}.}
  \bibinfo{year}{1992}\natexlab{a}.
\newblock \emph{\bibinfo{title}{Bayesian methods for adaptive models}}.
\newblock \bibinfo{thesistype}{Ph.D. Dissertation}. \bibinfo{school}{California
  Institute of Technology}.
\newblock


\bibitem[\protect\citeauthoryear{MacKay}{MacKay}{1992b}]%
        {mackay1992practical}
\bibfield{author}{\bibinfo{person}{David~JC MacKay}.}
  \bibinfo{year}{1992}\natexlab{b}.
\newblock \showarticletitle{A practical Bayesian framework for backpropagation
  networks}.
\newblock \bibinfo{journal}{\emph{Neural computation}} \bibinfo{volume}{4},
  \bibinfo{number}{3} (\bibinfo{year}{1992}), \bibinfo{pages}{448--472}.
\newblock


\bibitem[\protect\citeauthoryear{Malinin and Gales}{Malinin and Gales}{2018}]%
        {malinin2018predictive}
\bibfield{author}{\bibinfo{person}{Andrey Malinin} {and} \bibinfo{person}{Mark
  Gales}.} \bibinfo{year}{2018}\natexlab{}.
\newblock \showarticletitle{Predictive uncertainty estimation via prior
  networks}.
\newblock \bibinfo{journal}{\emph{arXiv preprint arXiv:1802.10501}}
  (\bibinfo{year}{2018}).
\newblock


\bibitem[\protect\citeauthoryear{Neal}{Neal}{1993}]%
        {neal1993bayesian}
\bibfield{author}{\bibinfo{person}{Radford~M Neal}.}
  \bibinfo{year}{1993}\natexlab{}.
\newblock \showarticletitle{Bayesian learning via stochastic dynamics}. In
  \bibinfo{booktitle}{\emph{Advances in neural information processing
  systems}}. \bibinfo{pages}{475--482}.
\newblock


\bibitem[\protect\citeauthoryear{Neal}{Neal}{2012}]%
        {neal2012bayesian}
\bibfield{author}{\bibinfo{person}{Radford~M Neal}.}
  \bibinfo{year}{2012}\natexlab{}.
\newblock \bibinfo{booktitle}{\emph{Bayesian learning for neural networks}}.
  Vol.~\bibinfo{volume}{118}.
\newblock \bibinfo{publisher}{Springer Science \& Business Media}.
\newblock


\bibitem[\protect\citeauthoryear{Netzer, Wang, Coates, Bissacco, Wu, and
  Ng}{Netzer et~al\mbox{.}}{2011}]%
        {netzer2011reading}
\bibfield{author}{\bibinfo{person}{Yuval Netzer}, \bibinfo{person}{Tao Wang},
  \bibinfo{person}{Adam Coates}, \bibinfo{person}{Alessandro Bissacco},
  \bibinfo{person}{Bo Wu}, {and} \bibinfo{person}{Andrew~Y Ng}.}
  \bibinfo{year}{2011}\natexlab{}.
\newblock \showarticletitle{Reading digits in natural images with unsupervised
  feature learning}.
\newblock  (\bibinfo{year}{2011}).
\newblock


\bibitem[\protect\citeauthoryear{Russakovsky, Deng, Su, Krause, Satheesh, Ma,
  Huang, Karpathy, Khosla, Bernstein, et~al\mbox{.}}{Russakovsky
  et~al\mbox{.}}{2015}]%
        {russakovsky2015imagenet}
\bibfield{author}{\bibinfo{person}{Olga Russakovsky}, \bibinfo{person}{Jia
  Deng}, \bibinfo{person}{Hao Su}, \bibinfo{person}{Jonathan Krause},
  \bibinfo{person}{Sanjeev Satheesh}, \bibinfo{person}{Sean Ma},
  \bibinfo{person}{Zhiheng Huang}, \bibinfo{person}{Andrej Karpathy},
  \bibinfo{person}{Aditya Khosla}, \bibinfo{person}{Michael Bernstein},
  {et~al\mbox{.}}} \bibinfo{year}{2015}\natexlab{}.
\newblock \showarticletitle{Imagenet large scale visual recognition challenge}.
\newblock \bibinfo{journal}{\emph{International journal of computer vision}}
  \bibinfo{volume}{115}, \bibinfo{number}{3} (\bibinfo{year}{2015}),
  \bibinfo{pages}{211--252}.
\newblock


\bibitem[\protect\citeauthoryear{Senge, B{\"o}sner, Dembczy{\'n}ski,
  Haasenritter, Hirsch, Donner-Banzhoff, and H{\"u}llermeier}{Senge
  et~al\mbox{.}}{2014}]%
        {senge2014reliable}
\bibfield{author}{\bibinfo{person}{Robin Senge}, \bibinfo{person}{Stefan
  B{\"o}sner}, \bibinfo{person}{Krzysztof Dembczy{\'n}ski},
  \bibinfo{person}{J{\"o}rg Haasenritter}, \bibinfo{person}{Oliver Hirsch},
  \bibinfo{person}{Norbert Donner-Banzhoff}, {and} \bibinfo{person}{Eyke
  H{\"u}llermeier}.} \bibinfo{year}{2014}\natexlab{}.
\newblock \showarticletitle{Reliable classification: Learning classifiers that
  distinguish aleatoric and epistemic uncertainty}.
\newblock \bibinfo{journal}{\emph{Information Sciences}}  \bibinfo{volume}{255}
  (\bibinfo{year}{2014}), \bibinfo{pages}{16--29}.
\newblock


\bibitem[\protect\citeauthoryear{Simonyan and Zisserman}{Simonyan and
  Zisserman}{2014}]%
        {simonyan2014very}
\bibfield{author}{\bibinfo{person}{Karen Simonyan} {and}
  \bibinfo{person}{Andrew Zisserman}.} \bibinfo{year}{2014}\natexlab{}.
\newblock \showarticletitle{Very deep convolutional networks for large-scale
  image recognition}.
\newblock \bibinfo{journal}{\emph{arXiv preprint arXiv:1409.1556}}
  (\bibinfo{year}{2014}).
\newblock


\bibitem[\protect\citeauthoryear{Srivastava, Hinton, Krizhevsky, Sutskever, and
  Salakhutdinov}{Srivastava et~al\mbox{.}}{2014}]%
        {srivastava2014dropout}
\bibfield{author}{\bibinfo{person}{Nitish Srivastava},
  \bibinfo{person}{Geoffrey Hinton}, \bibinfo{person}{Alex Krizhevsky},
  \bibinfo{person}{Ilya Sutskever}, {and} \bibinfo{person}{Ruslan
  Salakhutdinov}.} \bibinfo{year}{2014}\natexlab{}.
\newblock \showarticletitle{Dropout: a simple way to prevent neural networks
  from overfitting}.
\newblock \bibinfo{journal}{\emph{The journal of machine learning research}}
  \bibinfo{volume}{15}, \bibinfo{number}{1} (\bibinfo{year}{2014}),
  \bibinfo{pages}{1929--1958}.
\newblock


\bibitem[\protect\citeauthoryear{Wan, Zeiler, Zhang, Le~Cun, and Fergus}{Wan
  et~al\mbox{.}}{2013}]%
        {wan2013regularization}
\bibfield{author}{\bibinfo{person}{Li Wan}, \bibinfo{person}{Matthew Zeiler},
  \bibinfo{person}{Sixin Zhang}, \bibinfo{person}{Yann Le~Cun}, {and}
  \bibinfo{person}{Rob Fergus}.} \bibinfo{year}{2013}\natexlab{}.
\newblock \showarticletitle{Regularization of neural networks using
  dropconnect}. In \bibinfo{booktitle}{\emph{International conference on
  machine learning}}. PMLR, \bibinfo{pages}{1058--1066}.
\newblock


\bibitem[\protect\citeauthoryear{Wang, Li, Aertsen, Deprest, Ourselin, and
  Vercauteren}{Wang et~al\mbox{.}}{2019}]%
        {wang2019aleatoric}
\bibfield{author}{\bibinfo{person}{Guotai Wang}, \bibinfo{person}{Wenqi Li},
  \bibinfo{person}{Michael Aertsen}, \bibinfo{person}{Jan Deprest},
  \bibinfo{person}{S{\'e}bastien Ourselin}, {and} \bibinfo{person}{Tom
  Vercauteren}.} \bibinfo{year}{2019}\natexlab{}.
\newblock \showarticletitle{Aleatoric uncertainty estimation with test-time
  augmentation for medical image segmentation with convolutional neural
  networks}.
\newblock \bibinfo{journal}{\emph{Neurocomputing}}  \bibinfo{volume}{338}
  (\bibinfo{year}{2019}), \bibinfo{pages}{34--45}.
\newblock


\bibitem[\protect\citeauthoryear{Xie, Ma, Lei, Zhang, Xue, Tan, and Guo}{Xie
  et~al\mbox{.}}{2021}]%
        {9439951}
\bibfield{author}{\bibinfo{person}{Jiyang Xie}, \bibinfo{person}{Zhanyu Ma},
  \bibinfo{person}{Jianjun Lei}, \bibinfo{person}{Guoqiang Zhang},
  \bibinfo{person}{Jing-Hao Xue}, \bibinfo{person}{Zheng-Hua Tan}, {and}
  \bibinfo{person}{Jun Guo}.} \bibinfo{year}{2021}\natexlab{}.
\newblock \showarticletitle{Advanced Dropout: A Model-free Methodology for
  Bayesian Dropout Optimization}.
\newblock \bibinfo{journal}{\emph{IEEE Transactions on Pattern Analysis and
  Machine Intelligence}} (\bibinfo{year}{2021}), \bibinfo{pages}{1--1}.
\newblock
\urldef\tempurl%
\url{https://doi.org/10.1109/TPAMI.2021.3083089}
\showDOI{\tempurl}


\bibitem[\protect\citeauthoryear{Xie, Ma, Zhang, Xue, Chien, Lin, and Guo}{Xie
  et~al\mbox{.}}{2018}]%
        {xie2018balson}
\bibfield{author}{\bibinfo{person}{Jiyang Xie}, \bibinfo{person}{Zhanyu Ma},
  \bibinfo{person}{Guoqiang Zhang}, \bibinfo{person}{Jing-Hao Xue},
  \bibinfo{person}{Jen-Tzung Chien}, \bibinfo{person}{Zhiqing Lin}, {and}
  \bibinfo{person}{Jun Guo}.} \bibinfo{year}{2018}\natexlab{}.
\newblock \showarticletitle{Balson: Bayesian least squares optimization with
  nonnegative l1-norm constraint}. In \bibinfo{booktitle}{\emph{2018 IEEE 28th
  International Workshop on Machine Learning for Signal Processing (MLSP)}}.
  IEEE, \bibinfo{pages}{1--6}.
\newblock


\end{thebibliography}

\end{document}